\documentclass[10pt,twocolumn,letterpaper]{article}

\usepackage[pagenumbers]{cvpr} 

\definecolor{cvprblue}{rgb}{0.21,0.49,0.74}
\usepackage[pagebackref,breaklinks,colorlinks,allcolors=cvprblue]{hyperref}

\def\paperID{15} 
\def\confName{CVPR}
\def\confYear{2026}

\title{PointSplat: Efficient Geometry-Driven Pruning and Transformer Refinement for 3D Gaussian Splatting\\
}

\author{
Anh Thuan Tran, Jana Košecká\\
Department of Computer Science\\
George Mason University\\
{\tt\small \{atran92, kosecka\}@gmu.edu}\\ \\
\href{https://github.com/anhthuan1999/pointsplat}{github.com/anhthuan1999/pointsplat}
\vspace{-10pt}
}

\begin{document}
\maketitle
\begin{abstract}
3D Gaussian Splatting (3DGS) has recently unlocked real-time, high-fidelity novel view synthesis by representing scenes using explicit 3D primitives.
However, traditional methods often require millions of Gaussians to capture complex scenes, leading to significant memory and storage demands. Recent approaches have addressed this issue through pruning and per-scene fine-tuning of Gaussian parameters, thereby reducing the model size while maintaining visual quality. 
These strategies typically rely on 2D images to compute important scores followed by scene-specific optimization.
In this work, we introduce PointSplat, 3D geometry-driven prune-and-refine framework that bridges previously disjoint directions of gaussian pruning and transformer refinement. Our method includes two key components: (1) an efficient geometry-driven strategy that ranks Gaussians based solely on their 3D attributes, removing reliance on 2D images during pruning stage, and (2) a dual-branch encoder that separates, re-weights geometric and appearance to avoid feature imbalance. Extensive experiments on ScanNet++ and Replica across varying sparsity levels demonstrate that PointSplat consistently achieves competitive rendering quality and superior efficiency without additional per-scene optimization.
\end{abstract}

\section{Introduction}
\label{sec:intro}

\begin{figure}
  \centering
  \includegraphics[width=1\linewidth]{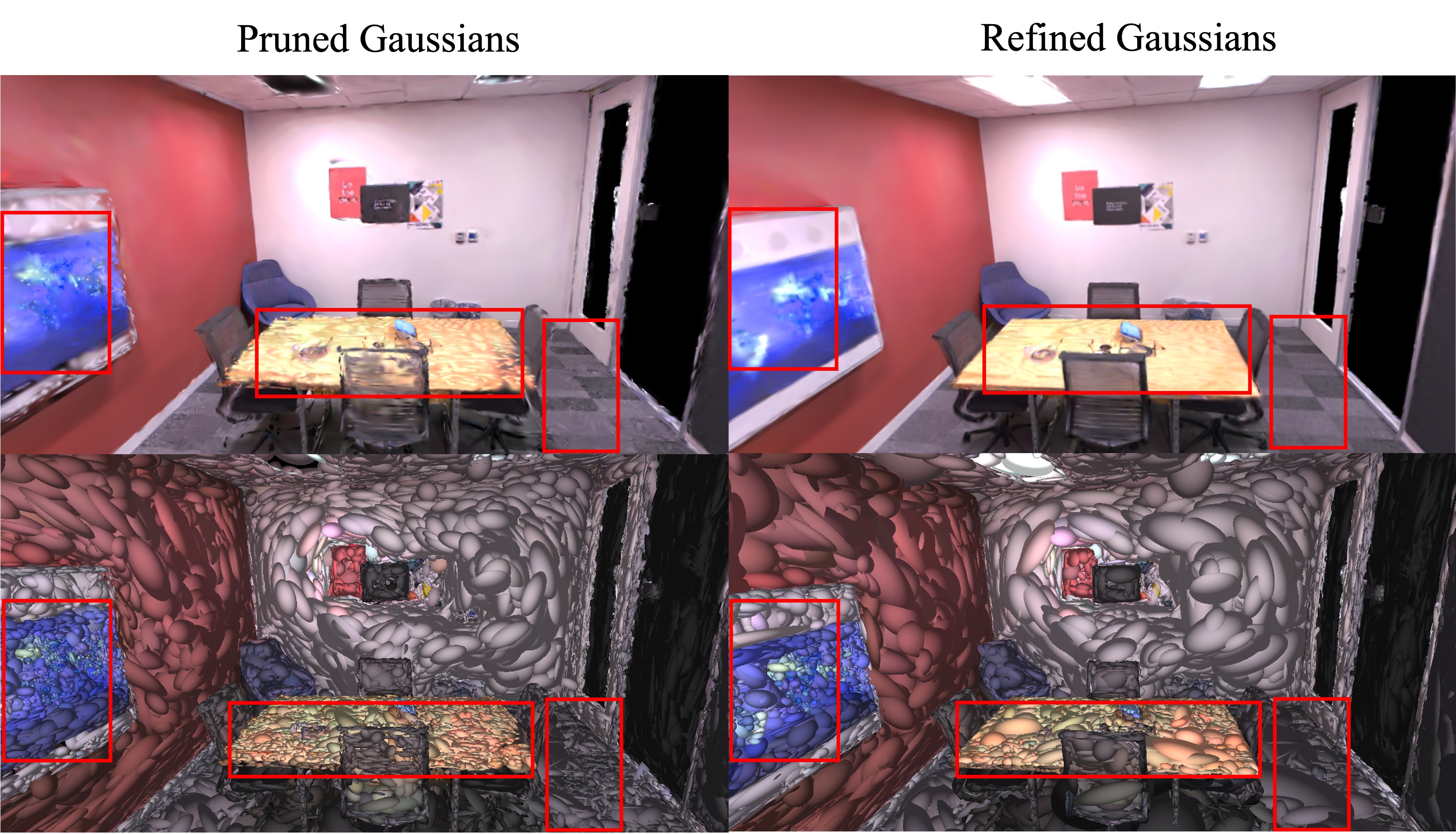}
  \caption{\textbf{Structural restoration through transformer-based refinement with rendered images (top) and 3D Gaussians (bottom).} While geometry-driven pruning (left) effectively reduces redundancy, it shows artifacts such as blurred textures and broken geometric edges. 3D transformer refinement (right) addresses these losses by globally coordinating Gaussian parameters to capture fine-grained structural details. This process provides compact, well-aligned ellipsoids that maintain high visual fidelity without the need for per-scene optimization.}
\end{figure}

Novel view synthesis (NVS) aims to render photorealistic images of 3D scenes from unseen viewpoints. This capability has broad applications across various fields, including virtual reality (VR), augmented reality (AR) \cite{3dgssurvey}, robotics \cite{3dgsrobotics}, and autonomous driving \cite{3dgsera}. Neural Radiance Fields (NeRF) \cite{nerf} reconstructs 3D representations from multiview images by mapping 3D coordinates and viewing directions to view-dependent color and volume density. The pixel intensities are computed through volume rendering techniques \cite{volumerendering}. While NeRF and its variants have achieved high-quality results, they suffer from slow inference speed, limiting their deployment in real-time or interactive applications.

Recent advances in 3D Gaussian Splatting (3DGS) \cite{3dgs} introduce a promising alternative with real-time rendering while maintaining competitive visual fidelity. 3DGS represented scenes through explicit 3D Gaussians, which are rendered using splatting techniques \cite{pointsplatting}. Specifically, each Gaussian is defined by position, scale, rotation, opacity, and coefficients of spherical harmonics modeling the viewpoint dependent appearance of the scene. While 3D Gaussians are typically initialized using Structure-from-Motion (SfM), the densification process and over-parameterization \cite{lightgaussian} can cause the number of Gaussians grow to millions per scene, resulting in significant memory demands and slow rendering performance. 
To address this issue, several studies have proposed methods to reduce the number of Gaussians through pruning based on importance scores. LightGaussians \cite{lightgaussian} quantifies Gaussian's contribution to each pixel ray across all training views, while  
Mini-Splatting \cite{mini} uses densification and simplification algorithm to reorganize the spatial positions of Gaussians rather than directly pruning them. 
PUP-3DGS \cite{pup3dgs} improves the scoring method using sensitivity-based pruning for higher compression ratios, while Speedy-Splat \cite{speedysplat} introduces hard and soft pruning algorithms to further accelerate rendering.
In prior approaches \cite{lightgaussian,pup3dgs,speedysplat}, pruning scores are typically computed using 2D images along with associated Gaussian attributes and corresponding camera rays, leading to expensive ray-pixel intersections.
Although these strategies improve rendering quality with fewer Gaussians, they need another per-scene optimization process after pruning stage. 
This dependency limits their practicality in downstream scenarios where obtaining test-time images for model compression or performing intensive per-scene tuning is computationally prohibitive, especially for real-time deployment on edge devices.

In this work, we propose PointSplat, 3D geometry-driven framework for efficient pruning and refinement of sparse 3D Gaussians. Our pruning score is derived solely from intrinsic 3D attributes, balancing spatial coverage and opacity to select a compact yet informative subset of Gaussians. This approach enables rapid compression by avoiding complex image-ray interactions. The retained Gaussians are then refined by Dual-Branch Encoder, which explicitly separates geometric attributes (position, scale, orientation) from high-dimensional appearance space (spherical harmonics) to avoid feature dominance with positional encoding for structured refinement. 
However, existing approaches to sparse 3DGS compression typically fall into two categories.
Image-driven pruning \cite{pup3dgs,lightgaussian,speedysplat} rely on 2D images and per-scene optimization, while transformer-only refinement \cite{splatformer} suffer from feature imbalance due to appearance dominance. 
PointSplat closes this gap by integrating geometry-driven pruning with transformer refinement, enabling high-quality reconstruction of compact Gaussians without requiring per-scene optimization at deployment.
Our main contributions can be described as follows:
\begin{itemize}
    \item We introduce an efficient geometry-driven pruning score based on Gaussian opacity and volume that selects informative primitives purely from 3D attributes, enabling rapid compression while preserving spatial coverage.
    \item We propose PointSplat network with Dual-Branch Encoder that decouples geometric information from high-dimensional appearance features to ensure stable and effective structural refinement.
    \item We conduct extensive experiments on ScanNet++ and Replica, demonstrating PointSplat consistently achieves competitive performance across multiple sparsity levels, validating its robustness and superior efficiency for deployment stage.
\end{itemize}

\section{Related Work}
\label{sec:formatting}
{Our work relates to improving the efficiency and practicality of pruning 3D Gaussian Splatting (3DGS) by addressing Gaussian redundancy and the limitations of existing pruning strategies,
that rely on 2D images and scene-specific optimization.} \\

\noindent
{\bf 3D Representations.} 
In Neural Radiance Fields (NeRF) \cite{nerf}, 3D scenes are represented using camera rays and multi-layer perceptrons (MLPs) to map coordinates and viewing directions to color and density values. This representation provides high-quality novel view synthesis through volume rendering along each ray. However, per-pixel MLP querying leads to slow inference speed, making it impractical for real-time applications.
To overcome speed limitation, several studies have proposed alternative spatial representations to accelerate NeRF-based rendering. TensorRF \cite{tensorf} models radiance fields as 4D tensors for efficient computation, while Instant-NGP \cite{instantngp} utilizes hash-based encoding to achieve faster convergence and real-time performance.
{To generalize on unseen scenes}, pixelNeRF \cite{pixelnerf} conditions radiance fields on sparse sets of input images and trained a single model for multiple scenes, while IBRNet \cite{ibrnet} uses a ray-transformer network that jointly encodes 3D locations and 2D viewing directions to synthesize novel views from few images.
Although these methods represented significant progress, achieving consistent real-time rendering, particularly for large-scale or complex scenes remains a challenge for NeRF-based approaches. Recent advances in point-based rendering via 3D Gaussian Splatting (3DGS) \cite{3dgs} have opened new directions for efficient novel view synthesis with competitive visual quality. 3DGS represents 3D scenes using large number (millions) of 3D Gaussians, each with position, orientation, scale, color, and opacity, followed by rendering novel views by directly projecting these Gaussians into image space and blending their colors \cite{pointsplatting}, enabling efficient real-time rendering.
To improve novel view synthesis from sparse set of views, pixelSplat \cite{pixelsplat} samples Gaussian means from dense probability distribution over 3D obtained by predefined encoder, while MVSplat \cite{mvsplat} localizes Gaussian centers using cost volume approach. {MVSGaussians \cite{mvsgaussian} further leverages multi-view stereo to encode point-wise features, which are then decoded into pixel-aligned Gaussian parameters, providing more effective initialization and faster fine-tuning.}
{Building on these advances, our method refines pretrained Gaussians by leveraging their explicit 3D geometric attributes. As a result, this approach allows efficient post-processing like pruning and refinement without scene-specific optimization.} \\

\noindent
{\bf Scalable and Progressive Representations.}
A parallel direction to pruning is the use of hierarchical or progressive Gaussians for adaptive streaming and level-of-detail (LoD).
Recent works such as PRoGS \cite{progs}, LapisGS \cite{lapisgs}, GoDe \cite{gode}, and LODGE \cite{lodge} build progressive LoD representations via either post-training or additional LoD construction with fine-tuning.
While effective, these approaches typically assume an explicit LoD construction pipeline for progressive rendering or streaming.
In contrast, PointSplat targets efficient pruning and refinement of existing overparameterized 3DGS models, allowing deployment-time compression without requiring specialized LoD training pipelines.

\noindent
{\bf Gaussian Ranking and Pruning.} 
While 3D Gaussian Splatting (3DGS) achieves remarkable rendering performance, it also suffers from redundancy in the number of Gaussians \cite{lightgaussian}, resulting in high memory demands. 
To address this, Compact-3DGS \cite{compact3dgs} introduces a codebook-based approach that groups similar Gaussians based on geometry, while LightGaussians \cite{lightgaussian} improves significance score calculation to include {global factors such as hit count from each pixel, along with vector quantization and distillation of spherical harmonics parameters}. PUP-3DGS \cite{pup3dgs} evaluates the sensitivity of each Gaussian to the rendered image and prunes accordingly. Mini-Splatting \cite{mini} improves the densification and simplification stages by assigning an importance score to each Gaussian during optimization. \textcolor{black}{In parallel, Speedy-Splat \cite{speedysplat} focuses on accelerating rendering through specialized rasterization kernels, while PointSplat operates at the representation level. Thus, PointSplat and Speedy-Splat are complementary rather than directly comparable.}
However, most existing pruning strategies rely on 2D images for score computation and require scene-specific fine-tuning. 
In contrast, PointSplat introduces a novel geometry-driven pruning score derived solely from intrinsic 3D attributes. By prioritizing spatially informative Gaussians and refining them through 3D transformer-based network, we eliminate the reliance on image-driven scoring and per-scene optimization at deployment.
\\

\noindent
{\bf Point Cloud Processing Refinements.} 
Recent advancements in 3D Gaussian Splatting (3DGS) have opened several opportunities \textcolor{black}{for using} point cloud understanding and processing architectures for further refinement of the Gaussian parameters. 
Instead of hand-crafted regularization techniques ~\cite{physgaussian}, several approaches adapted attention-based point transformer \cite{ptv1,ptv2,ptv3, pointct, stratified} to process the 3DGS as a point cloud set with Gaussian attributes serving
as features. SplatFormer \cite{splatformer} applied PointTransformerV3~\cite{ptv3} to refine 3D Gaussians, improving representation quality for out-of-distribution tasks. 
Large Spatial Model (LSM) \cite{lsm} used pretrained 2D transformer and 3D point-based transformer models to predict 3D Gaussians from two unposed input images. 

Previous studies face challenges when directly applying point-based transformer models to 3D Gaussians. Most point cloud transformer models have been developed for traditional point clouds, where each point typically contains only a small number of attributes (XYZ coordinates and RGB color). LSM \cite{lsm} addressed this by computing appearance through point-wise aggregation with color from stereo point maps, omitting high-dimensional spherical harmonics coefficients.
SplatFormer~\cite{splatformer} instead concatenates all Gaussian parameters, including opacity, scale, rotation, spherical harmonics, and position into a single feature vector and feeds it into PointTransformerV3 \cite{ptv3}. However, this approach suffered from feature imbalance: spherical harmonics coefficients can be up to 48 dimensions (for \(L=3\)), significantly outweighing other features like opacity (1D), scale, rotation, or position, leading to appearance-dominated representations that may lower effective learning. 
In contrast, PointSplat proposes Dual-Branch Encoder. This explicitly separates geometric and appearance features, reweights them with positional encoding, and prevents appearance dominance with stable learning.
\begin{figure*}
  \centering
  \includegraphics[width=1\linewidth]{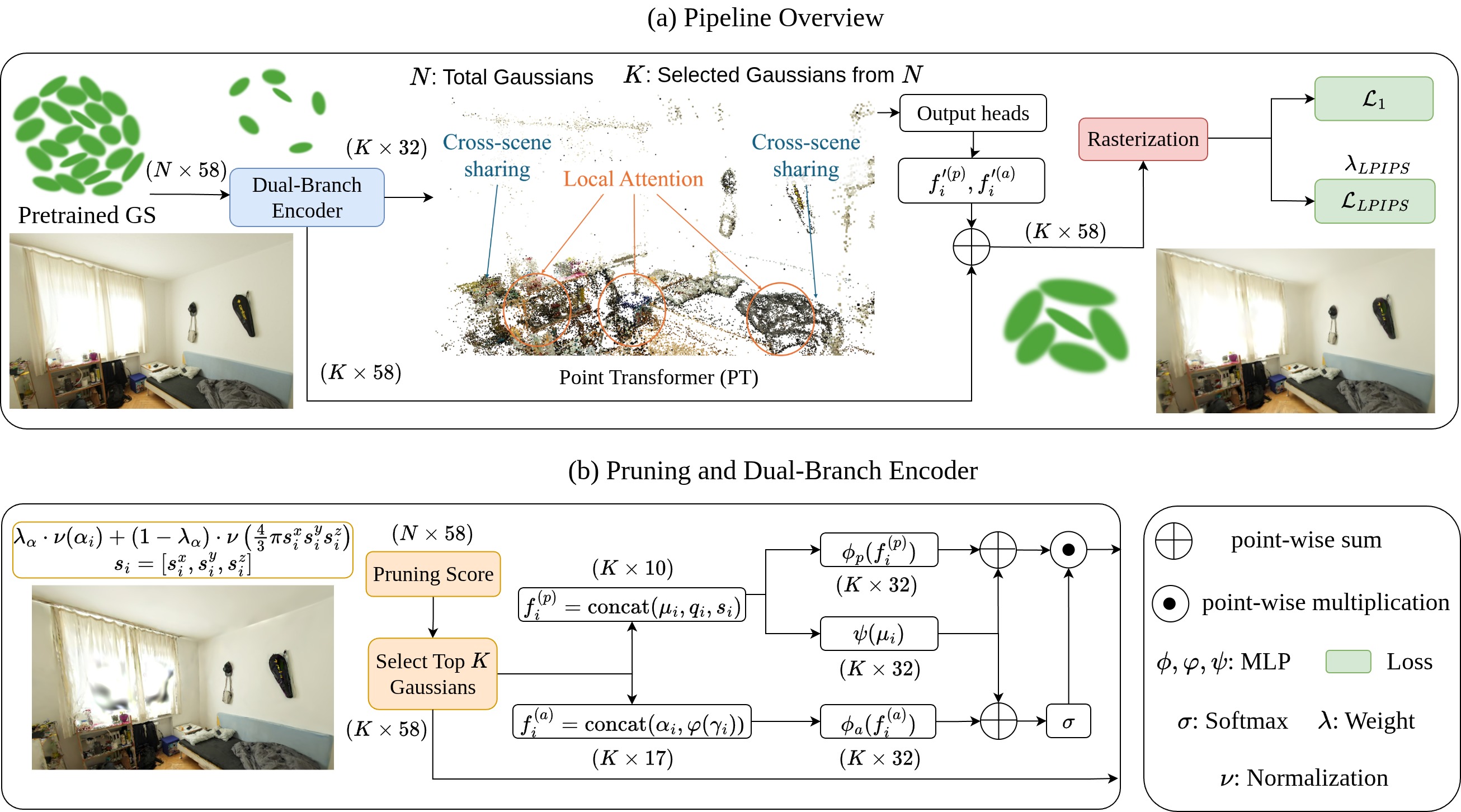}
  \caption{\textbf{Overall architecture.} (a) The pipeline begins with pretrained Gaussian Splatting (GS) model. Gaussians are first scored by the Dual-Branch Encoder (b), which selects an informative subset. Selected Gaussians are then encoded through separate geometry and appearance branches with positional encoding. These features are refined by Point Transformer and integrated with the pruned Gaussians through separate output heads. Finally, the sparse refined Gaussians are rasterized to render images and compute the training loss.
}
\end{figure*}

\section{Method}

In this work, we present PointSplat, a unified geometry-driven prune-and-refine framework.
Specifically, our approach addresses two key challenges: 
(1) selecting informative Gaussians that capture scene structure using only intrinsic 3D attributes, and 
(2) refining the retained Gaussians while balancing geometric and appearance features. 

\subsection{Preliminaries}

3D Gaussian Splatting (3DGS) \cite{3dgs} \textcolor{black}{represents} a scene using 3D Gaussians \(G = \{G_i\}_{i=1}^N\) where \(N\) is total number of Gaussians. Each Gaussian \(G_i\) is defined by its mean position \(\mu_i \in \mathbb{R}^3\) and covariance matrix \(\Sigma_i \in \mathbb{R}^{3\times3}\). Given 3D point $\mathbf x$, each Gaussian can be defined as:
\begin{equation}
G_i({\mathbf x}) \propto  \exp({-\frac{1}{2} ({\mathbf x}-{\mathbf{\mu}}_i)^{T} \Sigma_i^{-1}{({\mathbf x}-\mathbf{\mu}_i)}}) \quad 
\end{equation}
with covariance matrix \(\textstyle \Sigma_i = R_i S_i S_i^TR_i^T \), rotation matrix \(R_i\) parametrized by quaternion \(q_i \in \mathbb{R}^4 \), and diagonal scaling matrix \(S_i\) derived from scaling vector ${\mathbf s}_i = [s_i^x,s_i^y, s_i^z ]$. 
Each Gaussian has associated appearance attributes, including opacity value \(\alpha_i \in \mathbb{R}\) and color represented by coefficients of  spherical harmonics (SH) basis functions. Specifically, the color \(c_i\) is computed from SH coefficients \(\gamma(i) \in \mathbb{R}^{(L+1)^2 \times 3}\): 
\begin{equation}
    c_i = \sum_{l=0}^L \sum_{m=-l}^l \gamma^m_l(i)Y^m_l(\mathbf{v}_i)
\end{equation}
where \(L\) denotes SH degree, \({\mathbf v}_i\) is the viewing direction, and \(Y^m_l\) is the spherical harmonic basis function of degree \(l\) and order \(m\). While spherical harmonics of higher degree \(L\) can effectively capture view-dependent appearance, they also significantly increase the dimensionality of appearance attributes in 3DGS representations. At \(L=3\), each Gaussian requires 16  coefficients per color channel, totaling 48 coefficients. This increase in the model’s parameters can lead to an imbalance between the geometric (position) and appearance features of each Gaussian in the subsequent learning stage. 
During training~\cite{pointsplatting} parameters associated with the Gaussians are optimized end-to-end per scene using gradient descent to minimize a photometric loss over the input images. To improve rendering quality, densification stage splits, clones and prunes corresponding Gaussians. While this process increases reconstruction fidelity, it often expands the number of Gaussians from thousands to millions, leading to increased storage requirements.

\subsection{Geometry-Driven Pruning Score}

Our first objective is to select a subset of $K$ Gaussians  that preserves the global structure and visual appearance without relying on 2D images. While image-driven methods require expensive ray-pixel intersections, we propose an efficient geometry-driven pruning score that leverages intrinsic attributes. 

To balance geometric coverage and appearance contribution, the score is computed from Gaussian volume (derived from scale vector \(s_i\)) and opacity \(\alpha_i\). The weighted sum of these two terms is controlled by a trade-off parameter \(\lambda_\alpha\):
\begin{equation}
\mathrm{score}_i = \lambda_\alpha \cdot \nu(\alpha_i) + (1 - \lambda_\alpha) \cdot \nu\left(\frac{4}{3} \pi s_i^x s_i^y s_i^z\right)
\end{equation}
where  \(\nu(\alpha_i) \) and \(\nu(\frac{4}{3} \pi s_i^x s_i^y s_i^z)\) denote the z-score normalized opacity and volume of $i^{th}$ Gaussian and is computed for all \(N\) Gaussians. The weighting factor \(\lambda_\alpha \in [0, 1]\)  controls the balance between appearance and spatial contribution.
The top \(K\) Gaussians are selected based on this score,  prioritizing Gaussians that provide broad spatial coverage and contribute significantly to visual quality.
Since our refinement stage is built upon a transformer architecture operating on sparse 3D points, spatial coverage is a critical factor. A well-distributed selection of Gaussians ensures that the network receives enough information from diverse spatial regions, enabling it to effectively learn scene structure and appearance from the multiple images of complex indoor environments.

\subsection{Dual-Branch Encoder} 
Given sparse set \(K\) of Gaussians selected in the previous stage, we next describe how to refine their parameters.
Instead of using standard local gradient-based optimization for each scene, we use 3D point-based transformer ~\cite{ptv3} that globally 
coordinates the updates of Gaussians across the entire scene. 
While parameters in 3D Gaussian Splatting share some similarities with traditional 3D point clouds, they differ in several important ways. 
In various learning tasks on 3D point clouds, each point typically is defined by 3D coordinates (XYZ) and RGB color that are low-dimensional. Prior works \cite{ptv1, ptv2, ptv3} often concatenate these features directly as input to transformer-based architectures. In contrast, Gaussians in 3DGS have diverse and high-dimensional attributes, including 3D position \(\mu_i\), scale \(s_i\), rotation  \(q_i\)  (as quaternion), opacity  \(\alpha_i\), and spherical harmonics (SH) coefficients \(\gamma_i\) for view-dependent color representation. Among these, SH coefficients are particularly high-dimensional and simply concatenating all Gaussian parameters into a single feature vector, as done in SplatFormer \cite{splatformer}, the dominance of the SH coefficients can lead to feature imbalance. This can overshadow geometric factors, leading to biased feature learning and suboptimal model performance.

To address this, we introduce a dual-branch encoder for parameters associated with Gaussians that balances appearance and geometry related features before feeding them into the point transformer. The encoder consists of two parallel branches: geometric branch, which processes the 3D coordinates \(\mu_i\), scale \(s_i\), and rotation \(q_i\); and appearance branch, which processes the opacity \(\alpha_i\) together with reduced color features obtained by applying an MLP \(\varphi\) to the spherical harmonic coefficients, \(\varphi(\gamma_i)\).

\begin{equation}
f_i^{(p)} = \mathrm{concat}(\mu_i , q_i , s_i) \quad f_i^{(a)} = \mathrm{concat}(\alpha_i , \varphi(\gamma_i))
\end{equation}
\begin{equation}
\delta_i = \psi(\mu_i)
\end{equation}
\begin{equation}
f_i = \sigma(\phi_a(f_i^{(a)}) + \delta_i) \odot (\phi_p(f_i^{(p)}) + \delta_i)
\end{equation}

Each branch is passed through separate MLPs \(\phi_p, \phi_a\) to extract corresponding geometric and appearance features (see Figure 2b). Additionally, we introduce position encoding module to reinforce 3D spatial locations in input features. Unlike geometric branch, which includes intrinsic properties such as scale and rotation, the position encoding module \(\psi\) uses solely explicit 3D coordinates \(\mu_i\). 
This module ensures that we focus only the physical 3D locations of Gaussians when refining both geometric and appearance-related features, avoiding confusion with intrinsic Gaussian attributes, such as scale or rotation. The appearance features, combined with position encoding, are then transformed using softmax function \(\sigma\) yielding an attention weight for each Gaussian. 
These weights are applied to geometric features through point-wise multiplication, after adding position encoding to them as well.

The dual-branch encoder addresses two key challenges: (1) it mitigates the dominance of high-dimensional appearance features, preventing them from overwhelming geometric information; and (2) it enriches geometric features with appearance information and spatial location, resulting in more balanced and effective representations.\\
\textbf{3D Transformer Network.} 
Given the encoded features, we leverage 3D point transformer network \cite{ptv3} to optimize the appearance and geometric attributes of selected Gaussians. There are two core motivations behind using point transformer architecture in this work.
First, as illustrated in Figure 2a, point transformers employ local attention mechanisms, allowing each point to refine its features based on spatial neighbors. This locality constraint preserves fine-grained structural details while preventing influence from distant, unrelated Gaussians. 
Second, by operating on individual Gaussians with explicit 3D coordinates and enriched geometric features, the network can capture common structural characteristics across scenes, leading to more effective refinement of sparse Gaussians.

We treat the selected Gaussians as static 3D points, similar to point clouds, where each point is defined by its position and learned feature vector. 
These features are first processed through point transformer network (PT), and then passed through separate MLP output heads to get corresponding position and appearance Gaussian parameters. Finally, residual connections are applied to update the original Gaussian attributes before rasterization.

\begin{equation}
    f_i'^{(p)}, f_i'^{(a)} = \mathrm{PT}(f_i)
\end{equation}
\begin{equation}
    f_i^{(p)} = f_i^{(p)} + f_i'^{(p)}, \quad
    f_i^{(a)} = f_i^{(a)} + f_i'^{(a)}
\end{equation}

Furthermore, by learning features across multiple scenes, PointSplat network can refine sparse Gaussians that share similar properties (e.g., floors, walls), leading to better spatial coverage, as shown in Figure 1. This refinement improves scene quality and can be applied directly to pretrained 3DGS models without additional fine-tuning.

\section{Experimental Results}
\begin{table*}[t]
\centering
\caption{
Experimental results on ScanNet++ and Replica under varying Gaussian sparsity levels $G$ (\%). $G$ is the percentage of Gaussians remaining after pruning. 
Methods are categorized into \emph{Scene-Specific / Image-Driven} (require 2D images and per-scene optimization) and \emph{Geometry-Driven} (operate without images at deployment). \\
\textbf{\textit{*}} denote pruned Gaussians evaluated directly, without refinement through fine-tuning (scene-specific) or 3D networks (geometry-driven).
}

\begin{tabular}{clccc ccc}
\toprule
Remaining &  & \multicolumn{3}{c}{ScanNet++} & \multicolumn{3}{c}{Replica} \\
$G$ (\%) & Method & PSNR $\uparrow$ & SSIM $\uparrow$ & LPIPS $\downarrow$ & PSNR $\uparrow$ & SSIM $\uparrow$ & LPIPS $\downarrow$ \\ 
\midrule
100\% & 3DGS (Dense) & 29.73 & 0.925 & 0.165 & 39.40 & 0.974 & 0.122 \\
\midrule
\multicolumn{8}{l}{\textbf{Scene-Specific / Image-Driven Methods}} \\
\midrule 
50\% & LightGaussian* & 28.19 & 0.913 & 0.188 & 32.33 & 0.957 & 0.143 \\
   & LightGaussian & 30.23 & 0.929 & 0.171 & \textbf{39.78} & \textbf{0.974} & 0.125\\
   & PUP-3DGS* & 28.29 & 0.913 & 0.185 & 32.44 & 0.955 & 0.158 \\
   & PUP-3DGS & \textbf{31.35} & \textbf{0.937} & \textbf{0.159} & 39.67 & 0.973 & \textbf{0.121} \\
   \midrule
30\% & LightGaussian* & 26.13 & 0.891 & 0.215 & 26.86 & 0.921 & 0.182 \\
   & LightGaussian & 29.80 & 0.923 & 0.183 & 38.63 & \textbf{0.970} & 0.135 \\
   & PUP-3DGS* & 24.80 & 0.882 & 0.224 & 23.20 & 0.896 & 0.200 \\
   & PUP-3DGS & \textbf{30.85} & \textbf{0.931} & \textbf{0.172} & \textbf{38.64} & 0.968 & \textbf{0.130} \\
   \midrule
10\% & LightGaussian* & 21.87 & 0.836 & 0.278 & 20.29 & 0.837 & 0.290\\
   & LightGaussian & 28.12 & 0.902 & 0.224 & 33.37 & 0.937 & 0.195 \\
   & PUP-3DGS* & 19.82 & 0.803 & 0.305 & 17.34 & 0.784 & 0.322 \\
   & PUP-3DGS & \textbf{29.64} & \textbf{0.915} & \textbf{0.205} & \textbf{36.04} & \textbf{0.951} & \textbf{0.169} \\
\midrule
\multicolumn{8}{l}{\textbf{Geometry-Driven Methods}} \\
\midrule
50\% & SplatFormer  & 21.27 & 0.857 & 0.348  & 30.79 & 0.927 & 0.233 \\ 
   & PointSplat* & 28.81 & 0.923 & 0.266 & 33.97 & 0.956 & 0.213   \\
   & PointSplat* + SplatFormer & 29.03 & 0.924 & 0.264 & 34.63 & 0.952 & 0.192 \\
   & PointSplat & \textbf{29.52} & \textbf{0.928} & \textbf{0.258} & \textbf{35.98} & \textbf{0.958} & \textbf{0.191}  \\
   \midrule
30\% & SplatFormer & 11.66 & 0.691 & 0.509  & 29.14 & 0.903 & 0.263 \\ 
   & PointSplat* & 27.15 & 0.909 & \textbf{0.287} & 30.01 & 0.934 & 0.245 \\
   & PointSplat* + SplatFormer & 26.66 & 0.899 & 0.300 & 33.04 & 0.939 & \textbf{0.207} \\
   & PointSplat & \textbf{27.70} & \textbf{0.911} & \textbf{0.287} & \textbf{34.46} & \textbf{0.947} & 0.208 \\
   \midrule
10\% & SplatFormer & 10.62 & 0.691 & 0.569  & 8.37 & 0.388 & 0.677 \\
   & PointSplat* & 21.88 & 0.838 & 0.377 & 17.60 & 0.802 & 0.397 \\
   & PointSplat* + SplatFormer & 22.59 & 0.847 & 0.375 & 27.99 & 0.888 & \textbf{0.271} \\
   & PointSplat & \textbf{23.46} & \textbf{0.855} & \textbf{0.374} & \textbf{29.81} & \textbf{0.902} & 0.276 \\
\bottomrule
\end{tabular}
\end{table*}
\paragraph{Datasets.}
We evaluate the proposed network on ScanNet++ \cite{scannet++} using the novel view synthesis task, with 5 test scenes. For Replica \cite{replica}, we follow the setup in \cite{niceslam} and evaluate on 3 test scenes. In both datasets, following prior works \cite{3dgs, splatformer}, 10\% of the images in each scene are reserved for testing, while the remaining 90\% are used to obtain initial pretrained 3DGS models.
While our PointSplat network chooses only two random images per scene for refinement, similar to prior works \cite{pixelsplat, mvsplat}, it is built upon pretrained 3DGS models that are trained using all available views.
\\
{\bf Implementation details.}
We implement our architecture using PyTorch on NVIDIA A100 80GB. We begin by obtaining pretrained 3DGS models (Splatfacto) from NerfStudio \cite{nerfstudio}, which provide 3D Gaussian positions along with corresponding features, including opacity, scale, rotation, and spherical harmonics coefficients. 
\\
The proposed network is trained for 10,000 iterations using the Adam optimizer. The loss function is combined from $\mathcal{L}_{1}$ loss with a weighted $\mathcal{L}_{\mbox{\tiny LPIPS}}$ Learned Perceptual Image Patch Similarity (LPIPS) loss (\(\lambda_{\mbox{\tiny LPIPS}}=0.1\)), following \cite{splatformer}, and is defined as 
\(\mathcal{L} = \mathcal{L}_{1} + \lambda_{\mbox{\tiny LPIPS}}\mathcal{L}_{\mbox{\tiny LPIPS}}\).
The learning rate is initialized to 1e-5 and dropped by 10 at iteration 6,000. 
In the downsampling stages of the Point Transformer, we configure a 5-stage hierarchical architecture, with the number of transformer blocks in each stage set to (2, 2, 2, 4, 2), respectively.
Each MLP layer contains linear layers followed by batch normalization and ReLU activation. We fix \(\lambda_{\alpha} = 0.3\) for both datasets; results remain stable across a range of values, demonstrating robustness.
\\
{\bf Evaluation metrics.}
We follow standard practice of \cite{3dgs,splatformer} and report PSNR (Peak Signal-to-Noise Ratio), SSIM (Structured Similarity Index Metric), and LPIPS (Learned Perceptual Image Patch Similarity) metrics to evaluate rendering quality.
\subsection{Quantitative Results} 
The proposed geometry-driven pruning score alone provides significant improvements, boosting the performance of both 3DGS and SplatFormer under heavy pruning. As shown in Table 1, SplatFormer applied to Gaussians pruned with our score (\textit{PointSplat* + SplatFormer}) increases PSNR from 21.27 to 29.03 on ScanNet++ at 50\% sparsity, and from 8.37 to 27.99 on Replica at 10\%. 

Beyond pruning, our Dual-Branch Encoder effectively addresses feature imbalance in transformer-based network. While scene-specific methods can exceed PointSplat after per-scene refinement, our approach provides robust performance rapidly through transformer inference, highlighting its efficiency and geometry-driven nature. In contrast to LightGaussian and PUP-3DGS, which require recomputing scores for each scene, PointSplat applies the same learned refinement model to test scenes without retraining or per-scene adaptation.

While PointSplat shows slightly higher LPIPS than \textit{PointSplat* + SplatFormer} in some cases (0.276 vs. 0.271 on Replica at 10\%), these differences are marginal compared to its consistently superior PSNR (\(27.99\to29.81\)) and SSIM (\(0.888\to0.902\)). This trade-off shows that PointSplat prioritizes geometric fidelity, recovering sharper structures and cleaner edges while maintaining competitive perceptual quality.
As illustrated in Figure 1 and Table 1, neither pruning nor refinement can achieve optimal performance: pruning alone tends to produce broken geometry and artifacts, while transformer-only refinement suffers from feature imbalance. By refining compact, high-opacity Gaussians into structural representations, PointSplat provides a highly efficient yet expressive solution for sparse 3DGS.

\begin{figure}
  \centering
  \includegraphics[width=1\linewidth]{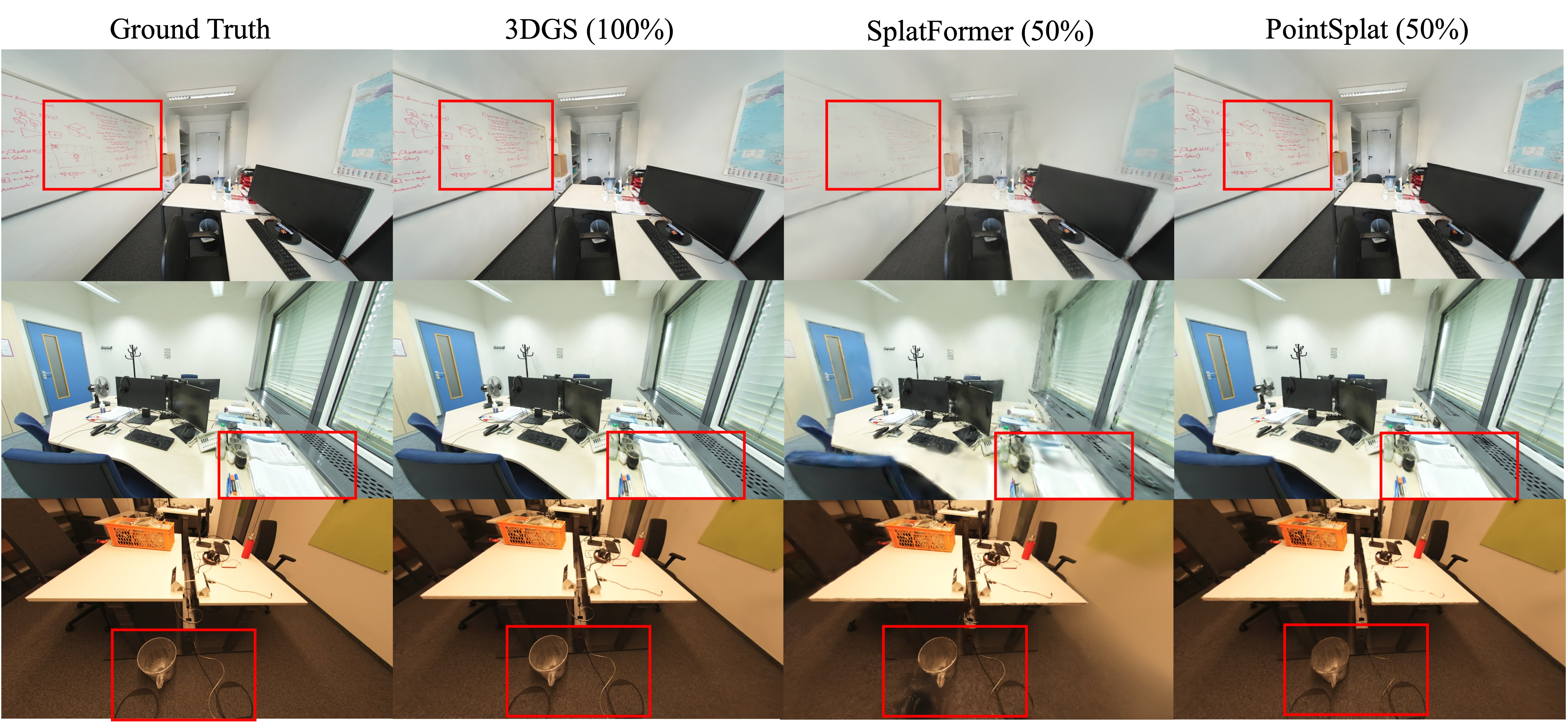}
  \caption{\textbf{Qualitative comparison.}  Ground truth images, pretrained dense 3DGS, and reconstructions from SplatFormer and PointSplat with 50\% of
  Gaussians. PointSplat delivers sharper details and more precise structures,
  preserves fine textures and complex shapes across scenes, achieving higher perceptual and structural fidelity without per-scene fine-tuning.}
\end{figure}
\subsection{Ablation study}
Ablation studies are conducted on 10\% sparsity level. 
\begin{table*}
  \centering
  \setlength{\tabcolsep}{5pt}
  \renewcommand{\arraystretch}{1.05}
  \begin{minipage}[t]{0.4\textwidth}
    \centering
    \small
    \caption{Efficiency results: $\#G$ = number of Gaussians.} 
    \label{tab:efficiency}
    \begin{tabular}{llccc}
      \toprule
      $G$ (\%) & Method & $\#G$ & FPS & Size(MB) \\
      \midrule
      100\% & 3DGS (Dense) & 588K & 207.41 & 145.92 \\
      \midrule
      50\%  & LightGaussian & 294K & 213.30 & 72.96 \\
          & PUP-3DGS     & 294K & 278.26 & 72.96 \\
          & PointSplat & 294K & \textbf{390.51} & 72.96 \\
      \midrule
      30\%  & LightGaussian & 176K & 253.42 & 43.78 \\
          & PUP-3DGS     & 176K & 352.71 & 43.78 \\
          & PointSplat & 176K & \textbf{440.56} & 43.78 \\
      \midrule
      10\%  & LightGaussian &  59K & 356.04 & 14.58 \\
          & PUP-3DGS     &  59K & 579.98 & 14.58 \\
          & PointSplat &  59K & \textbf{621.53} & 14.58 \\
      \bottomrule
    \end{tabular}
  \end{minipage}\hfill
  \begin{minipage}[t]{0.57\textwidth}
    \centering
    \scriptsize
    \caption{Ablation on $\lambda_\alpha$: PointSplat* denotes pruned Gaussians before refinement, and PointSplat after transformer refinement.}
    \resizebox{\linewidth}{!}{
    \begin{tabular}{c|c|ccc|ccc}
    \toprule
     & & \multicolumn{3}{c|}{ScanNet++} & \multicolumn{3}{c}{Replica} \\
    $\lambda_\alpha$ & Method & PSNR & SSIM & LPIPS & PSNR & SSIM & LPIPS \\
    \midrule
    0.1 & PointSplat* & 20.52 & 0.824 & 0.411 & 22.42 & 0.846 & 0.364 \\
        & PointSplat  & 21.72 & 0.837 & 0.406 & 28.23 & 0.881 & 0.281 \\
    0.3 & PointSplat* & 21.87 & 0.836 & 0.377 & 17.60 & 0.802 & 0.397 \\
        & PointSplat  & \textbf{23.46} & \textbf{0.855} & \textbf{0.374} & \textbf{29.81} & \textbf{0.902} & 0.276 \\
    0.5 & PointSplat* & 20.57 & 0.813 & 0.402 & 15.45 & 0.752 & 0.433 \\
        & PointSplat  & 22.72 & 0.846 & 0.376 & 27.71 & 0.888 & \textbf{0.274} \\
    0.7 & PointSplat* & 17.55 & 0.751 & 0.453 & 13.90 & 0.700 & 0.462 \\
        & PointSplat  & 20.10 & 0.816 & 0.410 & 27.39 & 0.886 & 0.281 \\
    0.9 & PointSplat* & 13.74 & 0.621 & 0.516 & 12.48 & 0.643 & 0.487 \\
        & PointSplat  & 10.66 & 0.686 & 0.559 & 26.58 & 0.882 & 0.290 \\
    \bottomrule
    \end{tabular}}
  \end{minipage}
\end{table*}
\noindent
{\bf Efficiency.}
\begin{figure}
  \centering
  \includegraphics[width=1\linewidth]{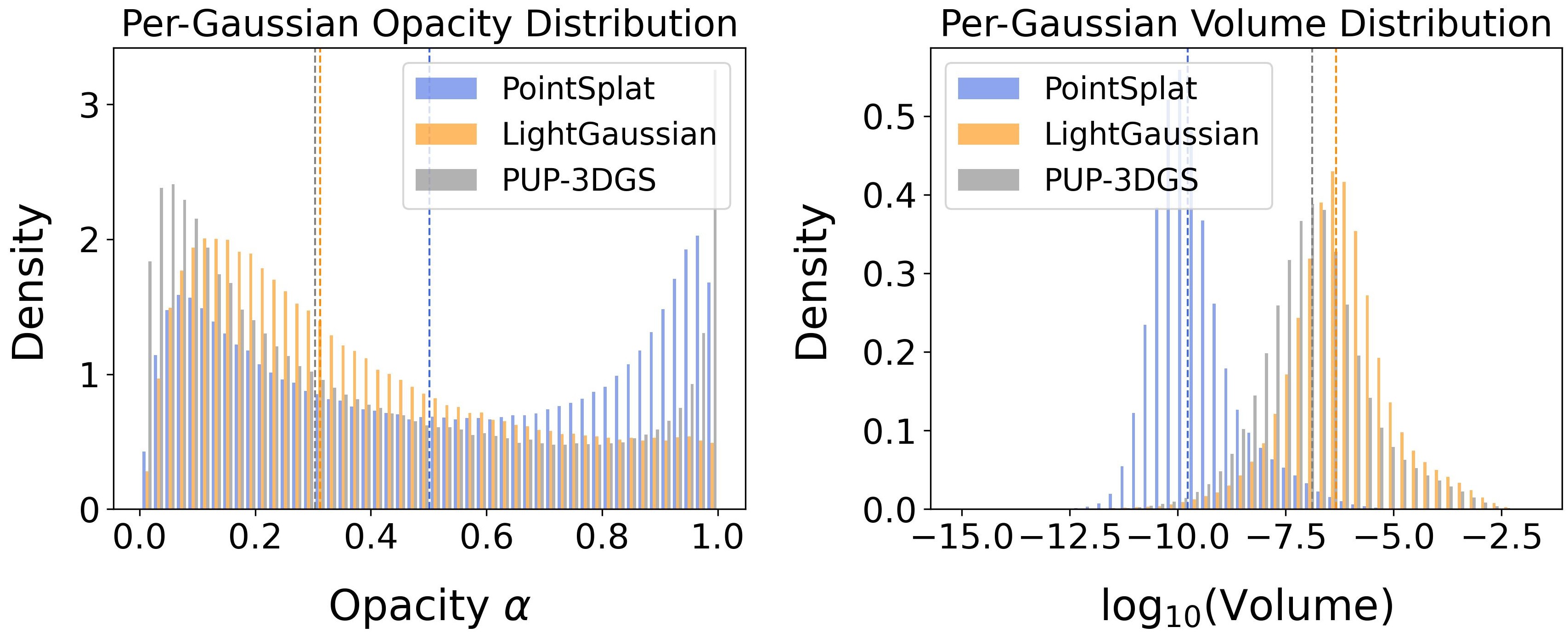}
  \caption{\textbf{Distribution of Gaussian opacity (left) and log-volume (right) at 50\% pruning.} Compared to LightGaussian (orange) and PUP-3DGS (gray), PointSplat (blue) concentrates on Gaussians with higher opacity and  smaller volumes. This contrasts with other studies, which leave large but weak splats. 
}
\end{figure}
Table 2 presents efficiency comparisons between PointSplat and prior image-driven pruning methods. 
At same number of Gaussians, PointSplat achieves consistently higher FPS than prior pruning methods at all sparsity levels. These results demonstrate the structural compactness of our representation. While traditional pruning often leaves large, low-opacity "floaters" that increase overdraw during rasterization, our geometry-driven score prioritizes smaller, high-opacity Gaussians. This reduces the number of primitives overlapping each pixel, thereby accelerating the splatting process.

To further analyze this efficiency boost, we show per-Gaussian statistics in Figure 4. The distributions show that PointSplat (blue) chooses smaller Gaussians (median $\log_{10}(\text{Volume}) \approx -10$) with higher opacity (median $\approx 0.5$). In contrast, image-driven baselines often retain larger, more transparent splats to fill missing structural details. \\
\noindent
{\bf Choosing Gaussians.}
Table 3 and Figure 5 show ablation study on pruning score weight \(\lambda_\alpha\), which controls the trade-off between appearance (opacity) and spatial coverage (Gaussian volume). We emphasize that \(\lambda_\alpha = 0.3\) is fixed globally across all datasets rather than tuned per dataset. 
This setting provides a balanced trade-off, thereby providing the highest PSNR and SSIM. Extremely low values (e.g., \(\lambda_\alpha = 0.1\)) overemphasize volume and select spatially large but visually weak Gaussians, while very high values (e.g., \(\lambda_\alpha = 0.9\)) overemphasize opacity and reduce spatial coverage, both of which degrade quality.
Moreover, PointSplat outperforms the baseline when selected Gaussians are informative, demonstrating its strong refinement capability. These results confirm that PointSplat remains robust across a range of \(\lambda_\alpha\) values and does not rely on sensitive hyperparameter tuning.
\begin{figure}
  \centering
  \includegraphics[width=1\linewidth]{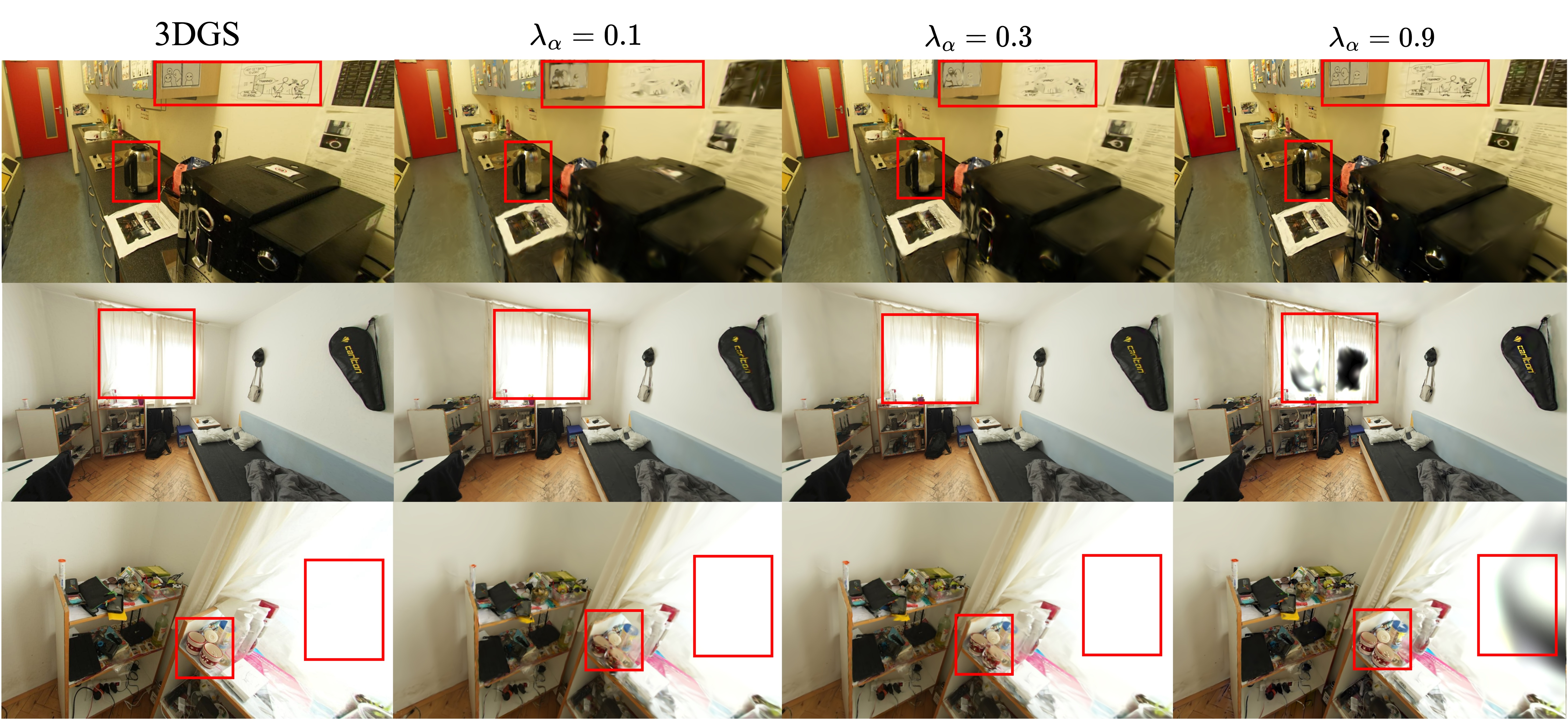}
  \caption{\textbf{Effect of pruning weight \(\lambda_\alpha\).}  
We visualize the pretrained 3DGS model before pruning and subsets selected with different \(\lambda_\alpha\).  
A balanced setting (\(\lambda_\alpha=0.3\)) maintains both spatial coverage and visual details.  
Lower values overemphasize volume and lose fine details, while higher values overemphasize opacity and produce uneven distributions.  
This demonstrates the need to balance geometry and appearance for robust sparse rendering.}
\end{figure}
\\
{\bf Dual-Branch Encoder.} 
Table 4 presents an ablation study of the proposed dual-branch encoder.
Model I serves as the baseline, 
using pruned 3DGS without refinement. Model II uses geometry features \(f_i^{(p)}\), such as scale and rotation.
Model III adds position encoding \(\delta_i\) to the geometry branch. Model IV uses appearance features \(f_i^{(a)}\) as weights applied to geometry features, but excludes position encoding. Finally, Model V uses both geometry and appearance features, each augmented with position encoding.
The results demonstrate that using geometry features alone (Model II) improves performance over the baseline (Model I). Adding position encoding to the geometry branch (Model III) provides further improvement, especially in SSIM and LPIPS. However, combining geometry and appearance without any positional encoding (Model IV) leads to weaker results, showing that spatial context is essential when fusing features. Model V achieves the best overall performance across all metrics, demonstrating the effectiveness of using geometry and appearance, both enhanced with position encoding, to improve spatial understanding. 
\begin{table}
\centering
\label{sample-table}
\centering
\caption{Ablation on the dual-branch encoder.}
{
\begin{tabular}{lcccccc}
\toprule
ID          & \(f_i^{(p)}\)     & \(f_i^{(a)}\) & \(\delta_i\) & PSNR & SSIM & LPIPS  \\
\midrule
I           &            &            &            & 21.87 & 0.836 & 0.377         \\
II          & \checkmark &            &            & 22.61 & 0.848 & 0.384     \\
III         & \checkmark &            & \checkmark & 22.76 & 0.847 & 0.385    \\
IV          & \checkmark & \checkmark &            & 22.65 & 0.846 & 0.388  \\
V           & \checkmark & \checkmark & \checkmark & \textbf{23.46} & \textbf{0.855} & \textbf{0.374}    \\
\bottomrule
\end{tabular}
}
\end{table}
\noindent \\\\
{\bf Limitation.} 
Despite its efficiency, PointSplat has certain limitations. First, our refinement transformer is used for indoor environments with similar structural properties. Extending to complex, unbounded outdoor scenes remains an open challenge. Second, while our pruning strategy is entirely geometry-driven, the training of transformer refinement still needs two images to ensure appearance consistency. 
\section{Conclusion}
By using pruning score derived solely from intrinsic 3D attributes, our method achieves rapid model compression without computationally expensive image-ray interaction.
To recover visual quality, we introduce Dual-Branch Encoder that mitigates feature imbalance and ensures stable learning of structural details.
Experiments on ScanNet++ and Replica show competitive rendering quality and superior efficiency of PointSplat compared to image-driven baselines. Overall, the resulting compact and expressive representations are well-suited for real-world applications in AR/VR and robotics, where reducing model size and accelerating deployment on resource-constrained devices are critical.\\
\noindent

\paragraph{Acknowledgement.}
This project was supported by resources provided by the Office of Research Computing at George Mason University (URL: https://orc.gmu.edu) and funded in part by grants from the National Science Foundation (Award Number 2018631).
{
    \small
    \bibliographystyle{ieeenat_fullname}
    \bibliography{main}
}


\end{document}